\documentclass[sigconf,10pt]{acmart}
\author[Huanqi Yang, Rucheng Wu, Weitao Xu]{Huanqi Yang, Rucheng Wu, Weitao Xu}
\affiliation{%
\institution{Department of Computer Science, City University of Hong Kong\\Email:\{huanqi.yang, ruchengwu2-c\}@my.cityu.edu.hk, weitaoxu@cityu.edu.hk}
\country{}
\
}

\copyrightyear{2024}
\acmYear{2024}
\setcopyright{acmlicensed}\acmConference[ACM MobiCom '24]{The 30th Annual
International Conference on Mobile Computing and Networking}{November 18--22,
2024}{Washington D.C., DC, USA}
\acmBooktitle{The 30th Annual International Conference on Mobile Computing and
Networking (ACM MobiCom '24), November 18--22, 2024, Washington D.C., DC, USA}
\acmDOI{10.1145/3636534.3698120}
\acmISBN{979-8-4007-0489-5/24/11}
\usepackage {color}
\usepackage{siunitx}
\usepackage{textcomp}
\usepackage{multirow}
\usepackage{graphicx}
\usepackage{geometry}
\usepackage{subfigure}
\usepackage{float}
\usepackage{url}
\usepackage{booktabs}
\hypersetup{
    colorlinks=true,
    linkcolor=red,
    urlcolor=magenta,
    citecolor=green,
}
\usepackage{siunitx}
\usepackage{bbding}
\usepackage{gensymb}
\usepackage{pifont}
\usepackage{arydshln}
\usepackage{diagbox}
\usepackage{gensymb}
\usepackage{fontawesome}
\usepackage{enumitem}
\usepackage{xspace}
\usepackage{sansmath}
\usepackage{makecell}
\usepackage[justification=centering]{caption}
\usepackage[ruled,linesnumbered]{algorithm2e}
\usepackage{tikz}
\usepackage{fp}
\usepackage{algpseudocode}
\usepackage{ragged2e}
\usepackage{mathrsfs}
\usepackage{amsthm}
\usepackage{multirow}
\usepackage{colortbl}
\usepackage{wasysym}
\usepackage{fontawesome}
\usepackage{xcolor}
\usepackage{threeparttable}
\newcommand{\SystemName}{\texttt{TransCompressor}\xspace}

\newlength\maxlentime

\def\headertime{New}
\definecolor{headerColor}{RGB}{173, 216, 230}
\definecolor{rowColor1}{RGB}{245, 245, 245}
\definecolor{rowColor2}{RGB}{224, 224, 224}

\begin{document}

\title{TransCompressor: LLM-Powered Multimodal Data Compression for Smart Transportation}

\renewcommand{\shorttitle}{TransCompressor}

\begin{abstract}
The incorporation of Large Language Models (LLMs) into smart transportation systems has paved the way for improving data management and operational efficiency. This study introduces \SystemName, a novel framework that leverages LLMs for efficient compression and decompression of multimodal transportation sensor data. \SystemName has undergone thorough evaluation with diverse sensor data types, including barometer, speed, and altitude measurements, across various transportation modes like buses, taxis, and Mass Transit Railways (MTRs). Comprehensive evaluation illustrates the effectiveness of \SystemName in reconstructing transportation sensor data at different compression ratios. The results highlight that, with well-crafted prompts, LLMs can utilize their vast knowledge base to contribute to data compression processes, enhancing data storage, analysis, and retrieval in smart transportation settings. 
\end{abstract}

% \begin{CCSXML}
% <ccs2012>
%    <concept>
%        <concept_id>10003120.10003138</concept_id>
%        <concept_desc>Human-centered computing~Ubiquitous and mobile computing</concept_desc>
%        <concept_significance>500</concept_significance>
%        </concept>
%  </ccs2012>
% \end{CCSXML}

% \ccsdesc[500]{Human-centered computing~Ubiquitous and mobile computing}

% \keywords{Data Compression, Large Language Models, Smart Transportation}

\maketitle
\settopmatter{printfolios=true}

\section{Introduction}
\label{sec:intro}
Recent developments in Large Language Models (LLMs) have demonstrated impressive capabilities~\cite{li2024personal,zhang2024large,yang2024drhouse}, as seen in applications like Sora~\cite{videoworldsimulators2024} and Claude-3~\cite{claude2023}. Sora showcases how LLMs can intuitively comprehend a broad spectrum of real-world phenomena, from fundamental physics principles to intricate societal and artistic concepts, mimicking a child's learning process. These LLMs not only excel in various tasks but also exhibit advanced human-like reasoning and exceptional generalization skills.
The potential of LLMs extends to the domain of smart transportation, where these models can enhance the intelligence of interconnected systems, enabling them to engage with and interpret the world in more sophisticated ways~\cite{zhang2024large}. This integration can lead to more efficient data processing, improved decision-making capabilities, and the ability to dynamically adapt to real-time changes in the environment, ultimately fostering smarter, more responsive transportation networks.

% Existing works integrating LLMs in transportation systems 
% has been widely studied. For example, Ren et al. [67] proposed TPLLM, a traffic prediction framework which leverages GPT-2 as the base LLM to provide embedding inputs for downstream tasks, including traffic flow prediction, and traffic missing data imputation. Also, in ST-LLM, a framework introduced by Liu et al. [52], a PFA LLM is utilized for training on traffic feature datasets and inferring on new data to produce intermediate results for the downstream regression task to perform spatial-temporal prediction. However, these methods rely on the training or fine-tuning of LLMs with specific structure. 
Previous research has extensively explored the integration of LLMs in transportation systems. For instance, TPLLM~\cite{ren2024tpllm} is a traffic prediction framework that employs GPT-2 as the foundational LLM to generate embedding inputs for tasks like traffic flow prediction and imputing missing traffic data. Similarly, in ST-LLM~\cite{liu2024spatial}, a PFA LLM is employed to train on traffic feature datasets and make inferences on new data to generate interim results for spatial-temporal prediction tasks. However, these approaches necessitate training or fine-tuning LLMs with specific architectures.
To address this issue, some researchers propose utilizing the zero-shot technique, which relies solely on instructions within prompts, thereby reducing the need for extensive training or fine-tuning. For instance, TrafficGPT~\cite{zhang2024trafficgpt}, a fusion of ChatGPT with traffic foundation models, showcases how zero-shot ChatGPT can analyze traffic data and offer valuable insights for transportation management systems. Additionally, UrbanGPT~\cite{li2024urbangpt} introduces a framework for urban traffic spatiotemporal prediction that harnesses the zero-shot reasoning capabilities of LLMs.
Modern LLMs have shown impressive performance in tasks generated on-the-fly without requiring fine-tuning~\cite{brown2020language}. The zero-shot technique can yield excellent task-agnostic performance. Without the necessity for training, LLMs can serve as valuable components in transportation systems.

\begin{figure}
\centering\includegraphics[width=0.8\linewidth]{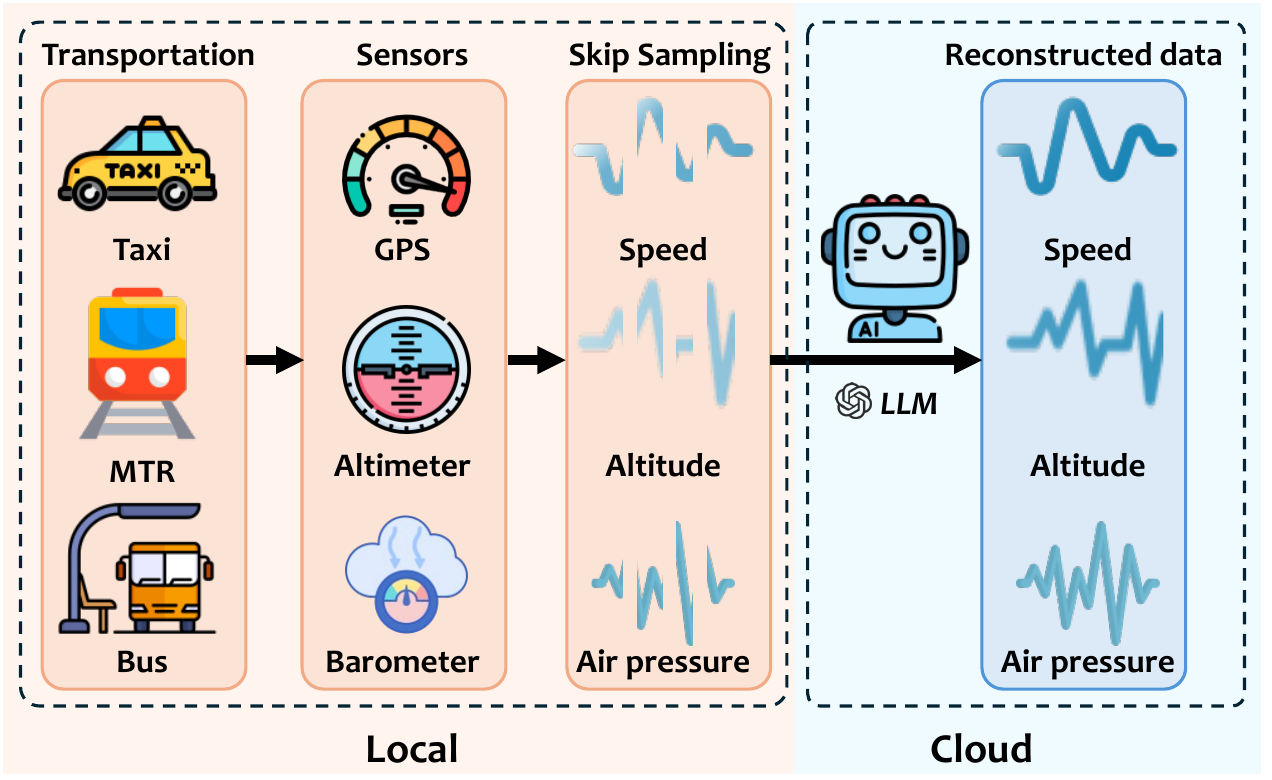}
    \caption{\textbf{Overview of \SystemName}}
    \label{fig:overview}
    \vspace{-0.2in}
\end{figure}

% Inspired by the advancements achieved with LLMs, this paper begins by centering on the application of sensor data compression in smart transportation systems. The aim is to explore and grasp the inferential and reconstructive capabilities of LLMs by feeding them compressed transportation sensor data along with a simple prompt. As illustrated in Fig.~\ref{fig:overview}, the skip sampled multimodal transportation sensor data is fed into LLM, leading to the models generating precise and effective reconstructions of the omitted data segments.
% Inspired by the advancements in this field, the paper proposes to use LLMs for sensor data compression in smart transportation systems. 
Inspired by the growing potential of LLMs to improve data management and operational efficiency in smart transportation systems, this paper proposes their use for sensor data compression. The objective is to investigate and understand the inferential and reconstructive capabilities of LLMs by providing them with compressed transportation sensor data along with a basic prompt. As depicted in Fig.~\ref{fig:overview}, the multimodal transportation sensor data is skip-sampled and sent to the cloud. The cloud then feeds this data into the LLM, enabling the generation of precise and efficient reconstructions of the original sensor data.
The performance evaluation covers three distinct scenarios: bus, taxi, and Mass Transit Railway (MTR), each incorporating three types of sensor data: barometer, speed, and altitude. The experimental results demonstrate that LLMs effectively execute zero-shot reconstruction with compressed transportation sensor data. \SystemName achieves a high reconstruction accuracy without necessitating retraining or fine-tuning for specific datasets.
The contribution can be summarized as:
\begin{itemize}
    \item We propose the first LLM-powered transportation sensor data compression scheme, which can enhance data transmission efficiency and reduce storage requirements in smart transportation systems.
    \item \SystemName shows that LLMs can serve as zero-shot data reconstructors, efficiently interpreting and decompressing data without requiring fine-tuning or training with domain-specific knowledge.
    \item Comprehensive evaluations show that our system can accurately reconstruct different types of transportation sensor data under various scenarios.
\end{itemize}
Additionally, we conduct a detailed analysis and discussion on the insights and significant discoveries derived from utilizing LLMs for transportation sensor data compression, laying the foundation for future research on integrating LLMs into smart transportation systems.

\section{Related Work}
\subsection{Large language models}
Recent advancements in LLMs have dramatically advanced artificial intelligence and natural language processing. LLMs such as OpenAI's GPT-3 and Meta's LLaMA 2~\cite{touvron2023llama} have shown exceptional capabilities in generating coherent and contextually relevant narratives, handling complex tasks such as multilingual translation, query responses, and code generation.
Historically, the development of neural language models (NLMs)\cite{arisoy2012deep} and early LLMs like GPT-2\cite{radford2019language} 
% and BERT~\cite{devlin2018bert} 
set foundational milestones. 
The development of these models has progressively featured enhanced complexity and capabilities, as exemplified by PaLM~\cite{chowdhery2023palm} and GPT-4.
Zero-shot generalization has significantly enhanced the utility of LLMs as assistants, prompting the development of methods aimed at aligning LLMs with human preferences and instructions. 
% In this study, we demonstrate that the zero-shot generalization capabilities of LLMs and their preference towards compressible patterns are not limited to language understanding but can also be effectively applied to transportation sensor data.

\subsection{LLMs for Time Series}

Several researchers have investigated the potential of using LLMs for time series applications. 
For example, TIME-LLM adapted an existing LLM for time series forecasting while preserving the original language model structure~\cite{jin2023time}. The model reprogrammed the input time series as text prototypes, facilitating alignment between text and time series modalities.
LLMTIME~\cite{gruver2024large} harnessed pretrained LLMs for continuous time series forecasting by representing numerical values in textual format and generating extrapolations through text completions. This model, however, only addresses the temporal dimension of the data, neglecting spatial aspects.
GATGPT combined the graph attention network with GPT for spatial-temporal imputation~\cite{chen2023gatgpt}, enhancing the LLM's ability to understand spatial dependencies, though it somewhat neglects the temporal aspects. Besides, some researchers propose to use LLMs to understand sensor data to sense the physical world~\cite{ji2024hargpt,yang2024you}. However, these methods, designed specifically for time series forecasting and understanding, are not suitable for transportation sensor data compression and reconstruction. Therefore, we propose \SystemName—a system that can effectively reconstruct compressed sensor data by leveraging the extensive knowledge base of LLMs.

\section{System Design}
This section introduces the design of \SystemName, which includes techniques such as skip sampling, data rescaling, and reconstruction. Specifically, skip sampling and data rescaling are performed on the transportation side, while reconstruction is carried out in the cloud as shown in Fig.~\ref{fig:overview}.

\subsection{Skip Sampling}
In modern transportation systems, particularly in sensor data collection, skip sampling is utilized as a method to reduce the volume of data without significantly compromising the integrity or utility of the data. This technique involves sampling only a specified fraction \( \alpha \) of the total data points collected. For example, if \( \alpha = 0.1 \), the system will sample every tenth data point.
The number of data points collected, \( n_{\text{collected}} \) of sequence \( x \) can be calculated as:
\[
n_{\text{collected}} = \lfloor \alpha \cdot n_{\text{total}} \rfloor,
\]
where \( n_{\text{total}} \) is the total number of data points that would have been collected without skip sampling, and \( \lfloor \cdot \rfloor \) denotes the floor function, which rounds down to the nearest whole number.
Skip sampling effectively decreases the burden on both the data storage systems and the bandwidth required for transmitting data to the cloud. This reduction is crucial for efficient and cost-effective system operations, especially in large-scale or real-time applications like traffic monitoring and autonomous vehicle navigation, where sending all data to the cloud in real-time may not be feasible.

\subsection{Data Rescaling}
Once the selectively sampled data reaches the cloud, it undergoes a process of rescaling. The goal of rescaling is to prepare the data for processing and analysis by ensuring that it conforms to a standard scale. Specifically, we rescale the data to the range between 0 and 1.
The formula for rescaling a data value \( x \) from its original range \([x_{\text{min}}, x_{\text{max}}]\) to the range \([0, 1]\) is given by:
\[
x_{\text{scaled}} = \frac{x - x_{\text{min}}}{x_{\text{max}} - x_{\text{min}}},
\]

After rescaling, the data undergoes a further process to retain only two decimal places. This truncation is performed to reduce the lengths of tokens, improving efficiency for LLMs. Truncated data is then ready to be fed into LLMs. The truncation can be mathematically represented as:
\[
x_{\text{truncated}} = \left\lfloor 100 \cdot x_{\text{scaled}} \right\rfloor / 100 ,
\]
where \( \left\lfloor \cdot \right\rfloor \) denotes the floor function, ensuring that the value is truncated to two decimal places.
\begin{figure}
    \centering
    \includegraphics[width=0.8\linewidth]{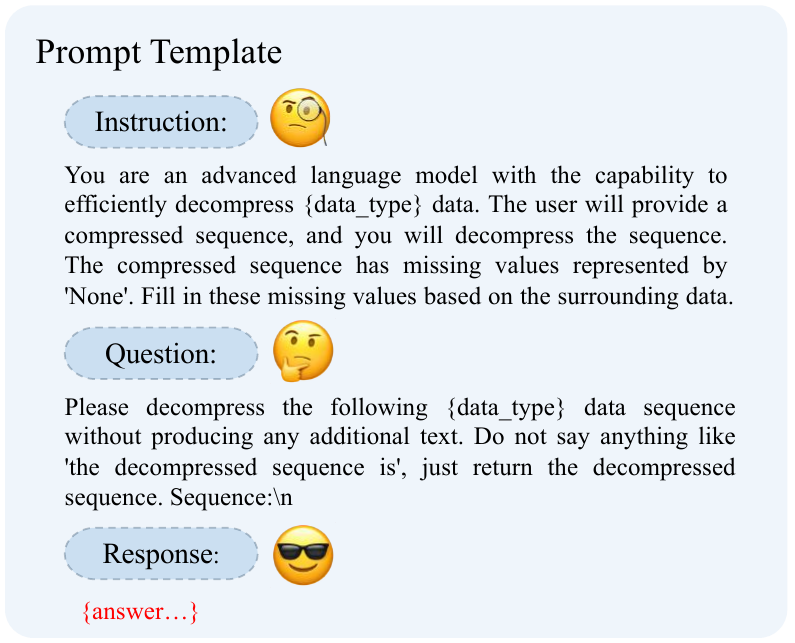}
    \caption{\textbf{Chain-of-thought prompt design for \SystemName.}}
    \label{fig:prompt}
    \vspace{-0.2in}
\end{figure}

\begin{figure*}
\centering
\subfigure[Bus altitude.]{
\begin{minipage}[t]{0.3\linewidth}
\centering
\includegraphics[width=\linewidth]{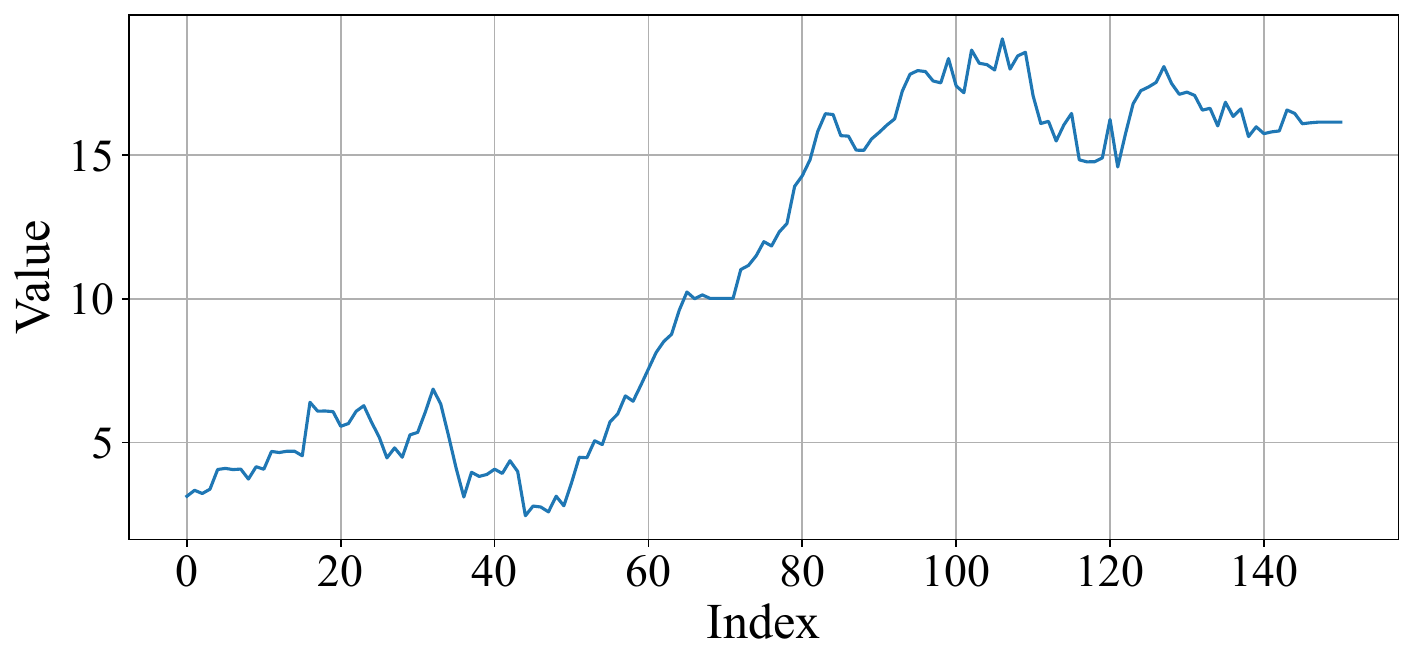}
\label{fig:bus1_altitude}
\end{minipage}%
}%
\subfigure[Bus pressure.]{
\begin{minipage}[t]{0.3\linewidth}
\centering
\includegraphics[width=\linewidth]{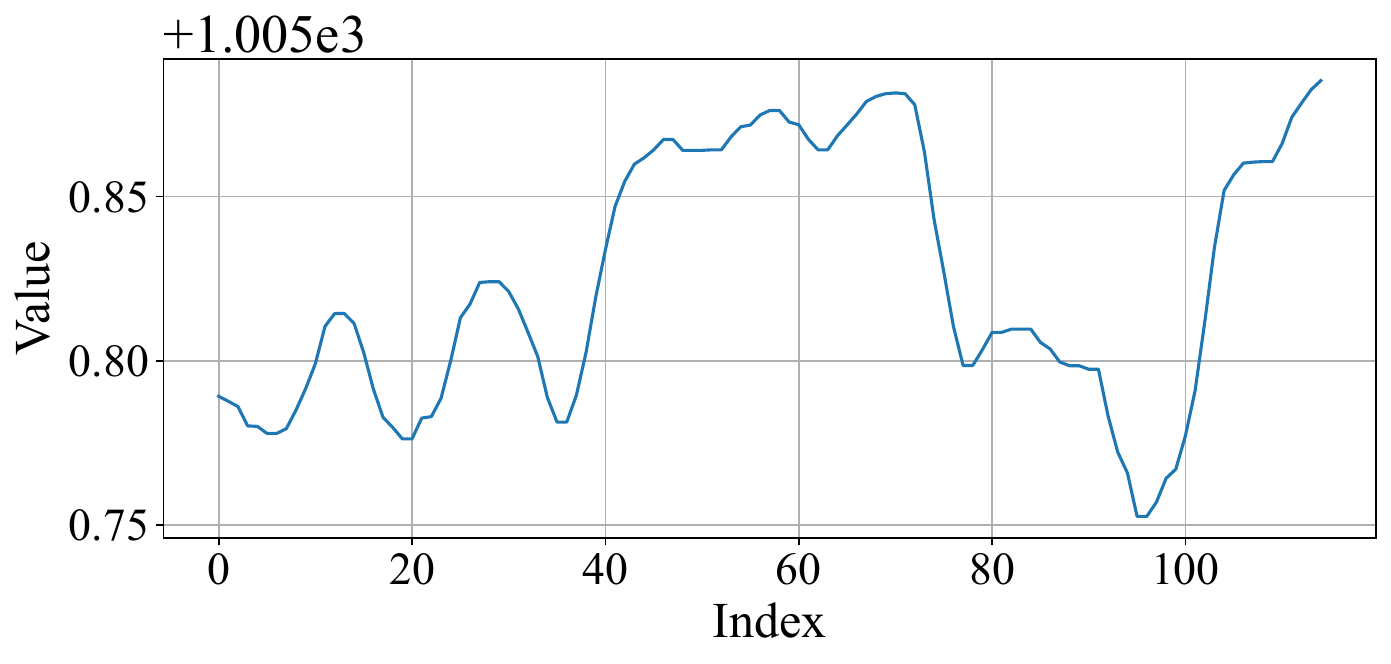}
\label{fig:bus1_pressure}
\end{minipage}%
}%
% \vspace{-3mm}
\subfigure[Bus speed.]{
\begin{minipage}[t]{0.3\linewidth}
\centering
\includegraphics[width=\linewidth]{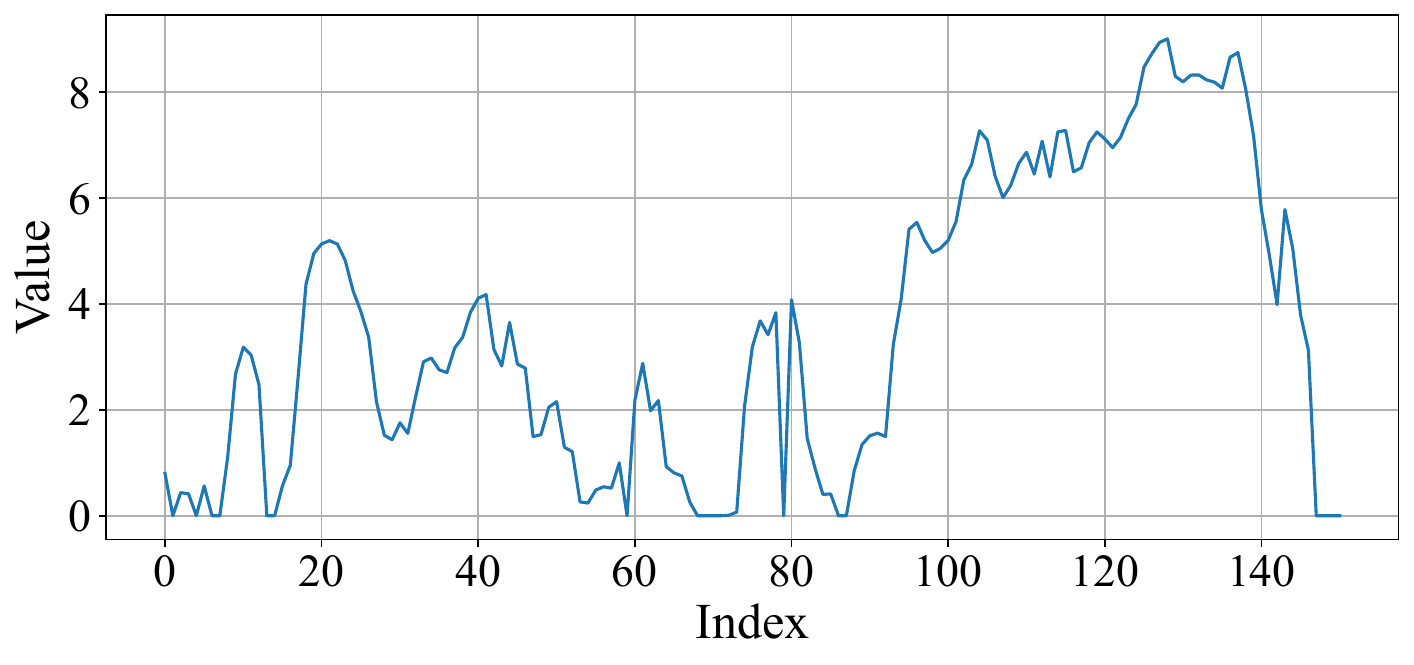}
\label{fig:bus1_speed}
\end{minipage}%
}%
\\
% \hspace{-5mm}
\subfigure[Taxi altitude.]{
\begin{minipage}[t]{0.3\linewidth}
\centering
\includegraphics[width=\linewidth]{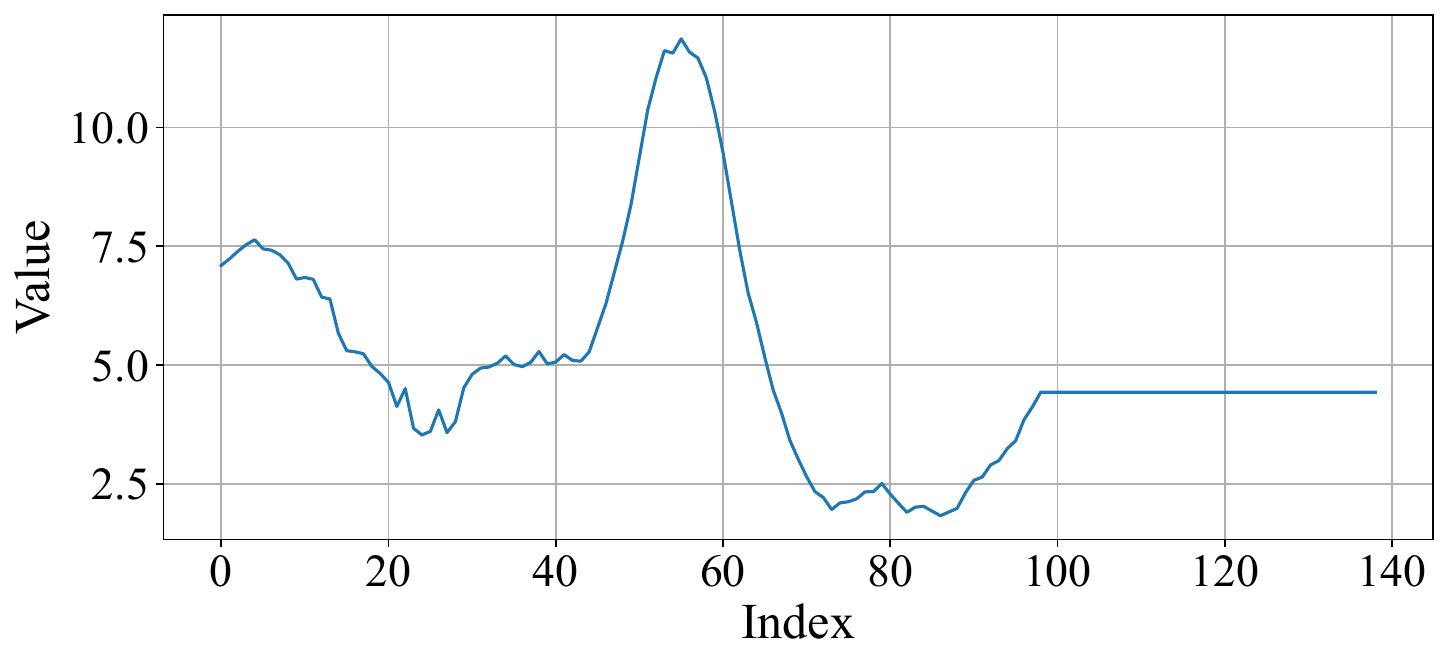}
\label{fig:taxi1_altitude}
\end{minipage}%
}%
\subfigure[Taxi pressure.]{
\begin{minipage}[t]{0.3\linewidth}
\centering
\includegraphics[width=\linewidth]{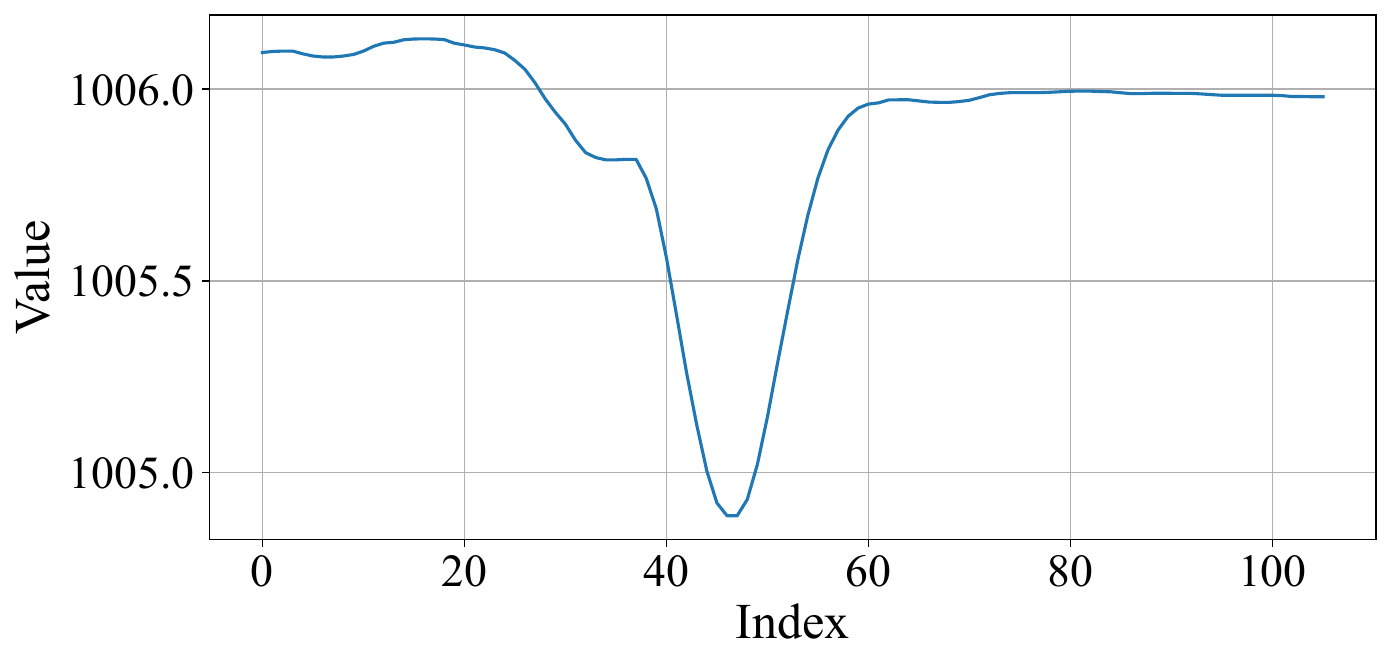}
\label{fig:taxi1_pressure}
\end{minipage}%
}%
\subfigure[Taxi speed.]{
\begin{minipage}[t]{0.3\linewidth}
\centering
\includegraphics[width=\linewidth]{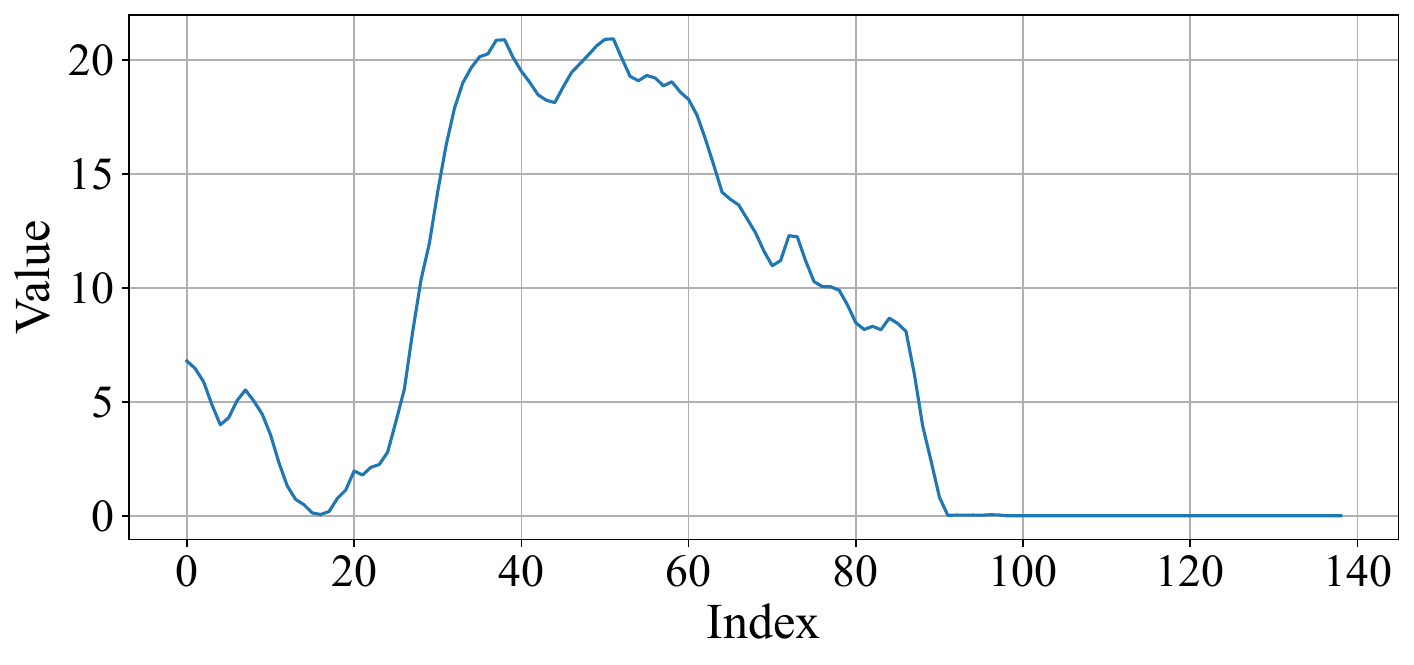}
\label{fig:taxi1_speed}
\end{minipage}%
}%
\centering
\caption{\textbf{Transportation sensor data.}}
\label{fig:sensor_data}
\vspace{-0.1in}
\end{figure*}

\subsection{Reconstruction.}
% The construction of our prompt, as visualized in Fig.~\ref{fig:prompt}, is elegantly minimalist, comprising solely of an initial instruction followed by a probing question.
% The directive is designed to activate the LLM's reservoir of specialized knowledge concerning transportation sensor data. Simultaneously, the question furnishes precise context about the data acquisition process, the downsampled sequence of transportation sensor data, and the sensor type being investigated.
% Our prompt culminates with an invitation for the LLM to "Do not say anything like
% 'the decompressed sequence is', just return the decompressed
% sequence." a tactic intended to obtain a clean result. 
Once the data has been transmitted to the cloud, we employ an LLM to reconstruct the sensor data.
The relationship between the input data and the LLM's output can be mathematically expressed as follows:
\[
y = \text{LLM}(x_{\text{truncated}}),
\]
where \( x_{\text{truncated}} \) represents the truncated version of the data, prepared and formatted as described in the previous sections. The function \( \text{LLM} \) encapsulates the processing capabilities of the Large Language Model, which computes the output \( y \) based on the input data.
The structure of our prompt, as illustrated in Fig.~\ref{fig:prompt}, adopts an elegantly minimalist design. It starts with a straightforward instruction that primes the LLM to tap into its specialized repository of knowledge on transportation sensor data. This is followed by a probing question that provides detailed context regarding the data acquisition process, the sequence of transportation sensor data, and the specific type of sensor under examination.
The prompt concludes with a directive to the LLM: "Do not say anything like 'the decompressed sequence is', just return the decompressed sequence." This approach is strategically chosen to ensure the output is clear and uncluttered, facilitating straightforward interpretation and analysis.
% This method is championed by existing research~\cite{kim2024health,wei2022chain}, for its efficacy in refining the precision of LLM outputs.
% As illustrated in Fig.~\ref{fig:gpt4inference}, by foregoing a restricted response format, we empower the LLM to freely generate a richer textual output. This approach leverages the model's inherent analytical capabilities to tap into and utilize the pertinent knowledge it has internalized.

\section{Evaluation}
\subsection{Experiment Setup}
\textbf{Data collection.} To effectively evaluate our system, we conducted real-world data collection across three different transportation modes: taxis, MTR, and buses. We utilized smartphones to capture multimodal sensor data, which included measurements of air pressure, speed, and altitude, from within these transportation vehicles. The smartphones were securely mounted within each vehicle to ensure stability and consistency in data gathering.
For each mode of transportation, we gathered a set of 30 data segments, with each segment consisting of 30 seconds of sensor readings. The sampling rate was standardized at 1Hz. Fig.~\ref{fig:sensor_data} illustrates a segment of the collected transportation sensor data.
% This methodology allowed us to acquire a comprehensive dataset that is representative of typical operational conditions across these transportation modes.
% , providing a robust basis for subsequent analysis and evaluation of our system.

\begin{figure*}
\centering
\subfigure[Taxi.]{
\begin{minipage}[t]{0.3\linewidth}
\centering
\includegraphics[width=\linewidth]{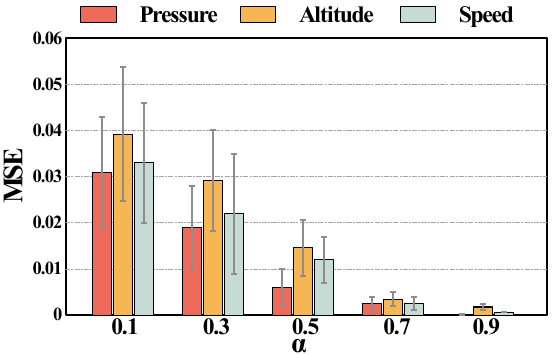}
\label{fig:eva1}
\end{minipage}%
}%
\subfigure[MTR.]{
\begin{minipage}[t]{0.3\linewidth}
\centering
\includegraphics[width=\linewidth]{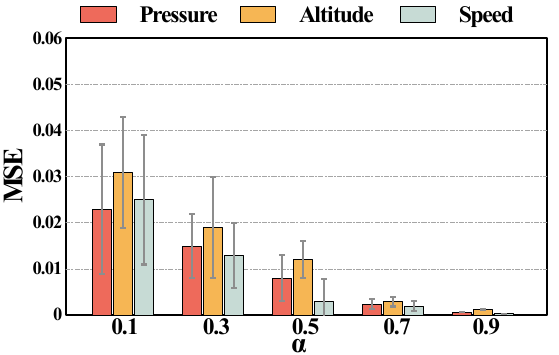}
\label{fig:eva2}
\end{minipage}%
}%
\subfigure[Bus.]{
\begin{minipage}[t]{0.3\linewidth}
\centering
\includegraphics[width=\linewidth]{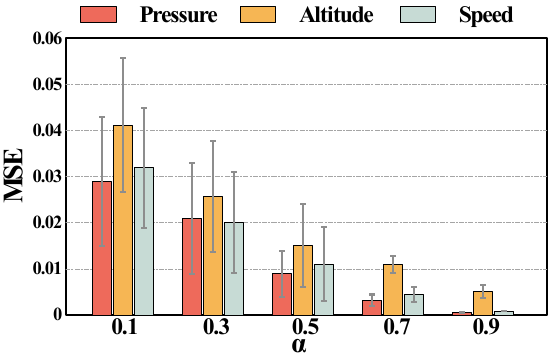}
\label{fig:eva3}
\end{minipage}%
}%
\centering
\caption{\textbf{Overall performance.}}
\label{fig:eva}
\vspace{-0.2in}
\end{figure*}

\begin{figure*}
\centering
% \setcounter{figure}{7}
% \subfigure[$\alpha$ = 0.3, air pressure.]{
% \begin{minipage}[t]{0.32\linewidth}
% \centering
% \includegraphics[width=\linewidth]{fig/Barometer_0.3_Window_1.pdf}
% \label{fig:r1}
% \end{minipage}%
% }%
% \subfigure[$\alpha$ = 0.3, speed.]{
% \begin{minipage}[t]{0.32\linewidth}
% \centering
% \includegraphics[width=\linewidth]{fig/speed_0.3_Window_1.pdf}
% \label{fig:r2}
% \end{minipage}%
% }%
% % \vspace{-3mm}
% \subfigure[$\alpha$ = 0.3, altitude.]{
% \begin{minipage}[t]{0.32\linewidth}
% \centering
% \includegraphics[width=\linewidth]{fig/altitude_0.3_Window_2.pdf}
% \label{fig:r3}
% \end{minipage}%
% }%
% \\
\subfigure[$\alpha$ = 0.5, air pressure.]{
\begin{minipage}[t]{0.3\linewidth}
\centering
\includegraphics[width=\linewidth]{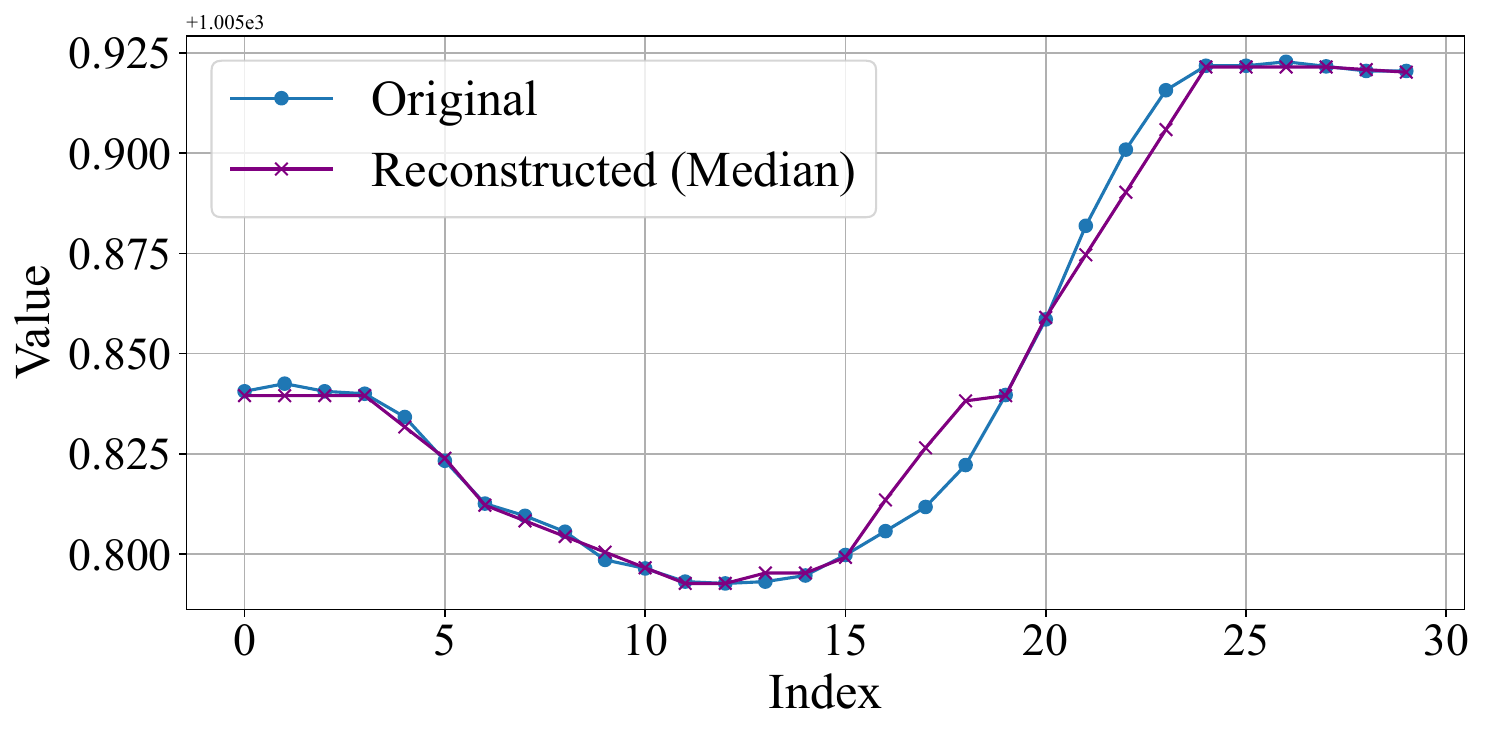}
\label{fig:r4}
\end{minipage}%
}%
\subfigure[$\alpha$ = 0.5, speed.]{
\begin{minipage}[t]{0.3\linewidth}
\centering
\includegraphics[width=\linewidth]{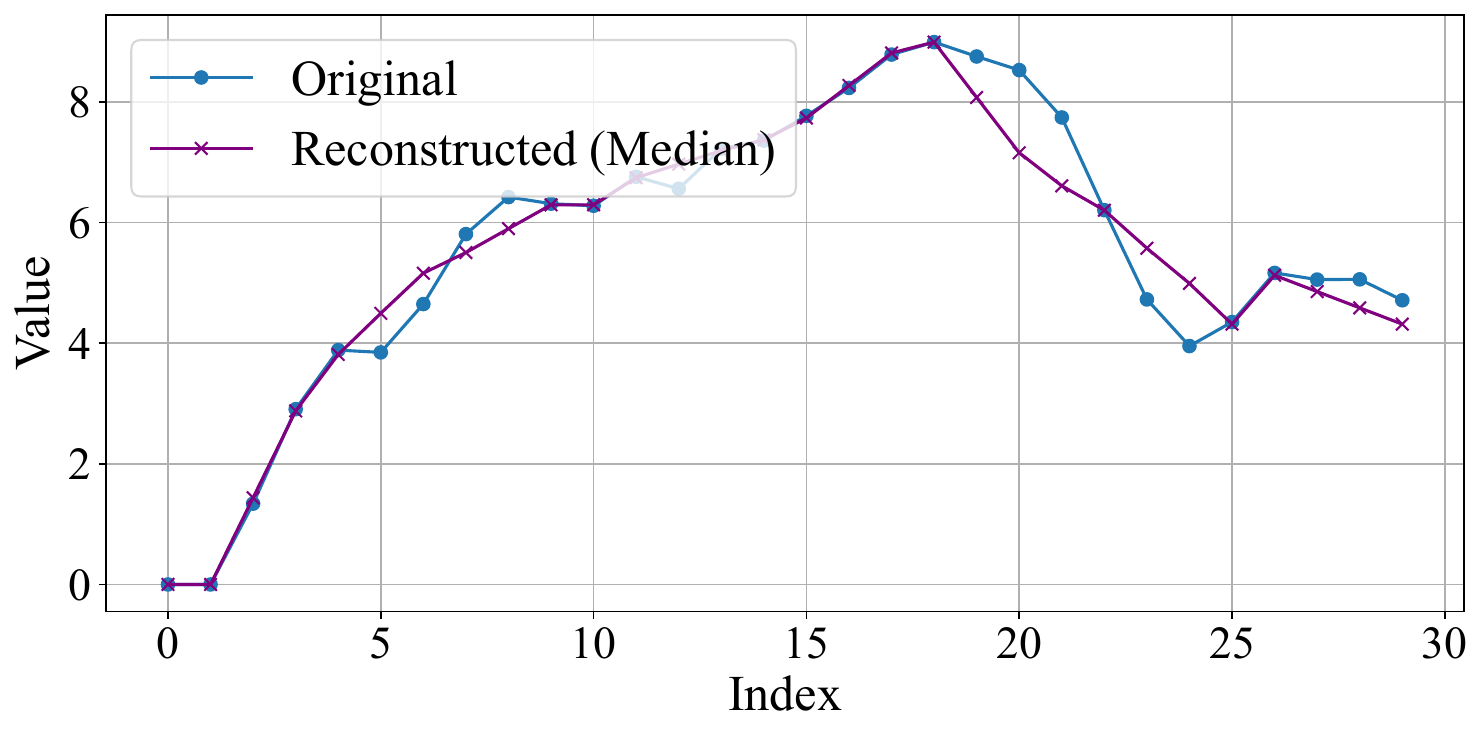}
\label{fig:r5}
\end{minipage}%
}%
\subfigure[$\alpha$ = 0.5, altitude.]{
\begin{minipage}[t]{0.3\linewidth}
\centering
\includegraphics[width=\linewidth]{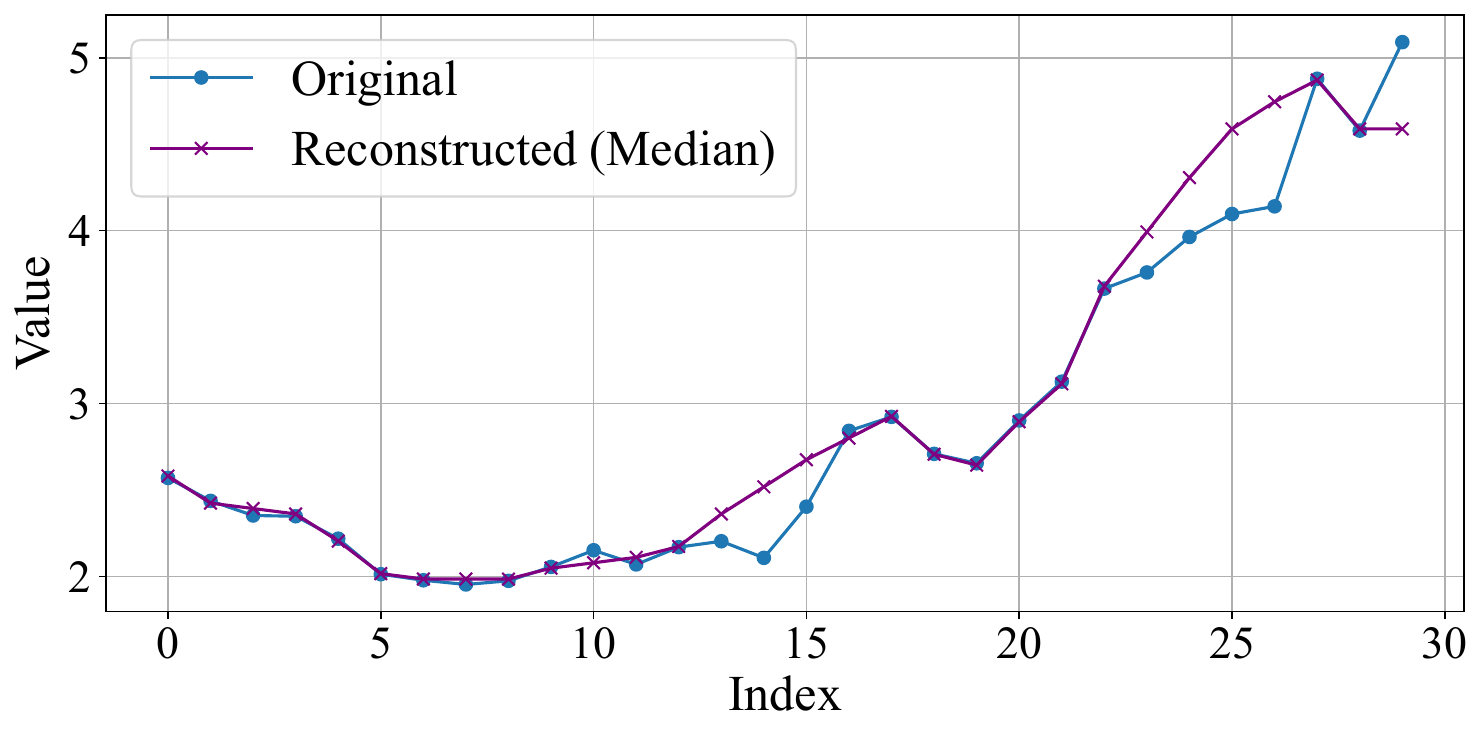}
\label{fig:r6}
\end{minipage}%
}%
\\
\subfigure[$\alpha$ = 0.7, air pressure.]{
\begin{minipage}[t]{0.3\linewidth}
\centering
\includegraphics[width=\linewidth]{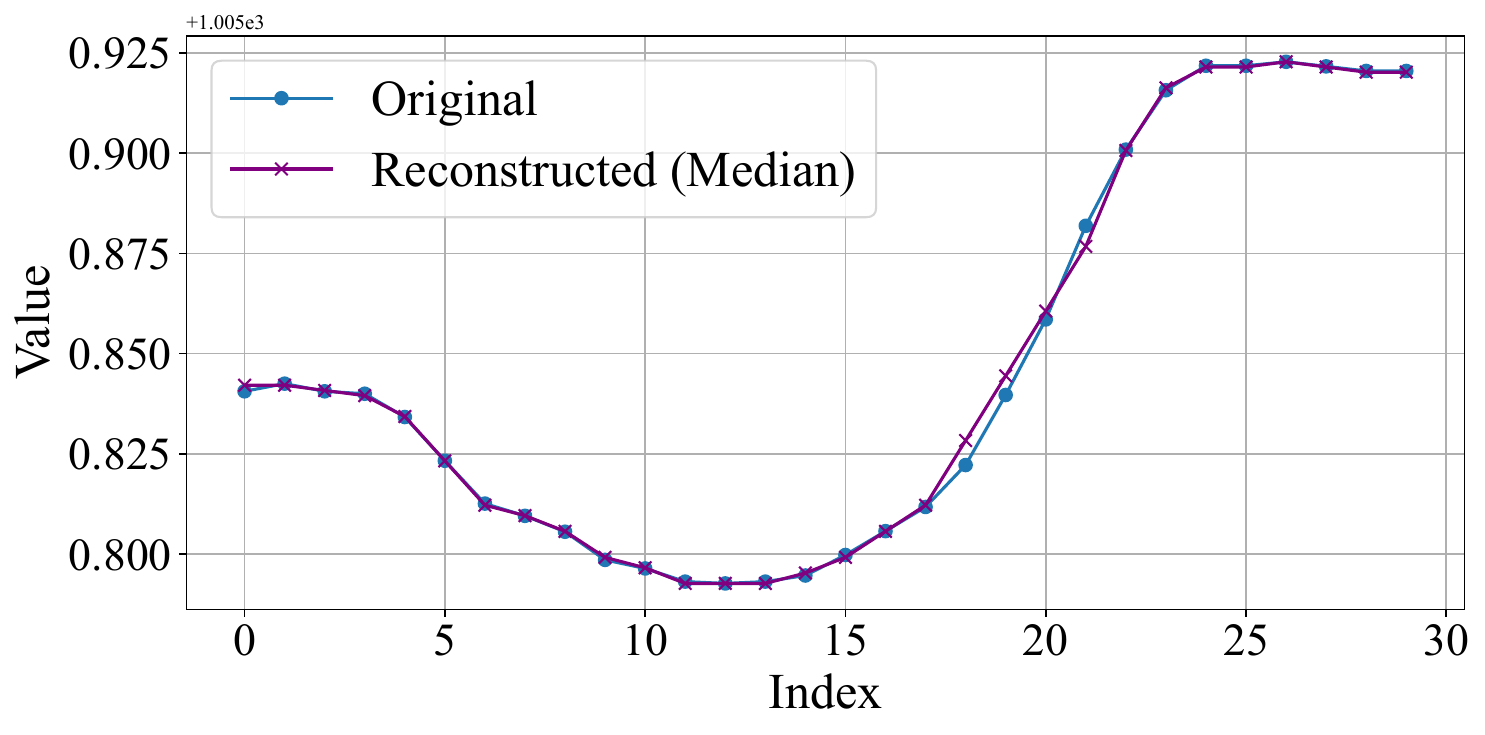}
\label{fig:r7}
\end{minipage}%
}%
\subfigure[$\alpha$ = 0.7, speed.]{
\begin{minipage}[t]{0.3\linewidth}
\centering
\includegraphics[width=\linewidth]{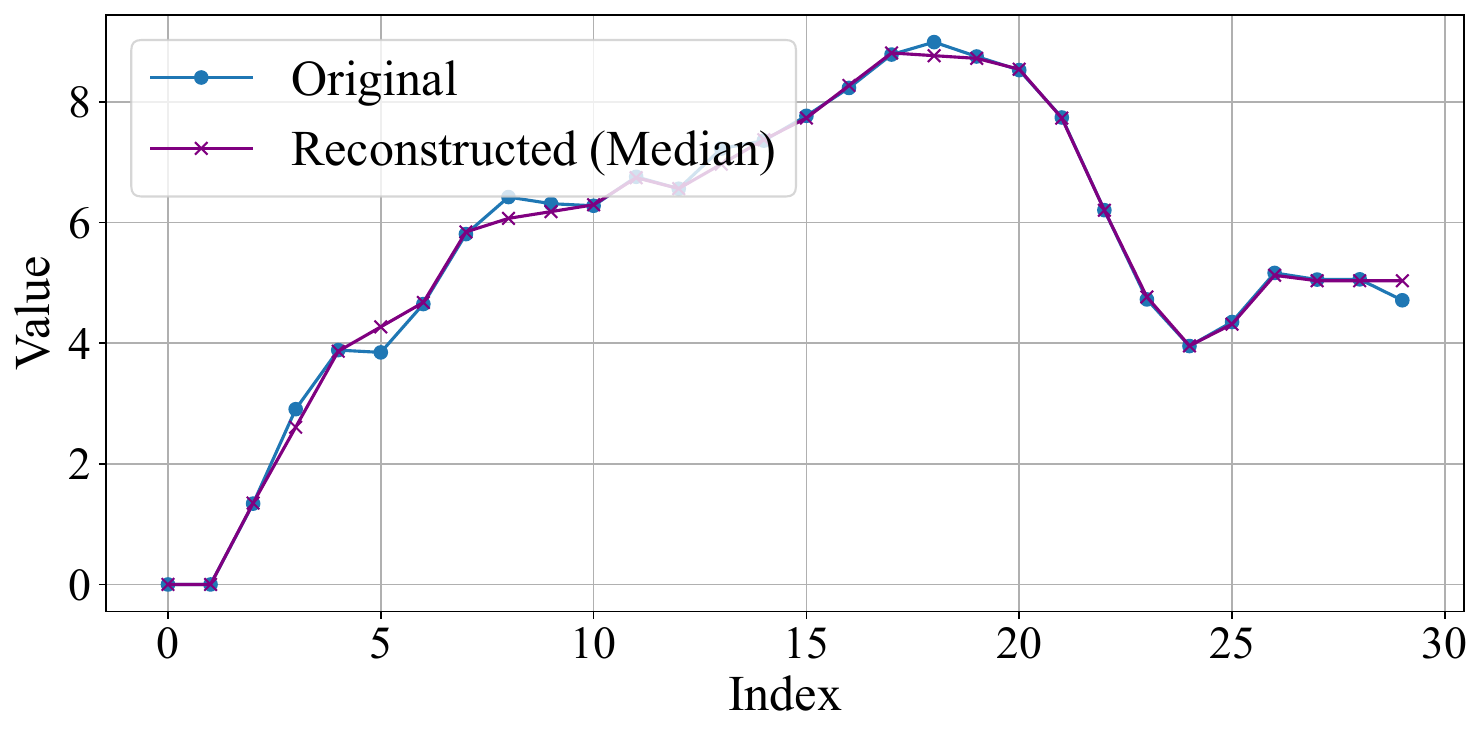}
\label{fig:r8}
\end{minipage}%
}%
% \vspace{-3mm}
\subfigure[$\alpha$ = 0.7, altitude.]{
\begin{minipage}[t]{0.3\linewidth}
\centering
\includegraphics[width=\linewidth]{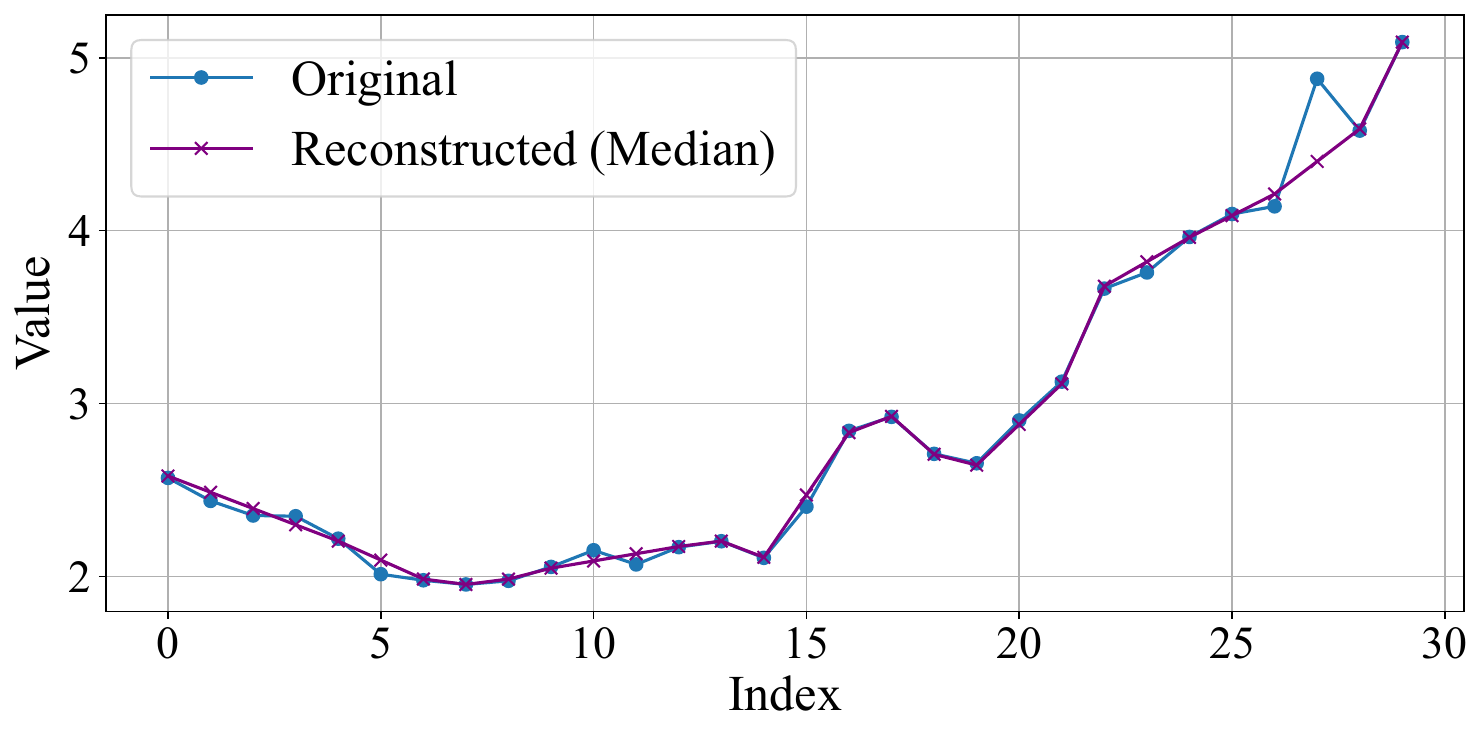}
\label{fig:r9}
\end{minipage}%
}%
\\
\subfigure[$\alpha$ = 0.9, air pressure.]{
\begin{minipage}[t]{0.3\linewidth}
\centering
\includegraphics[width=\linewidth]{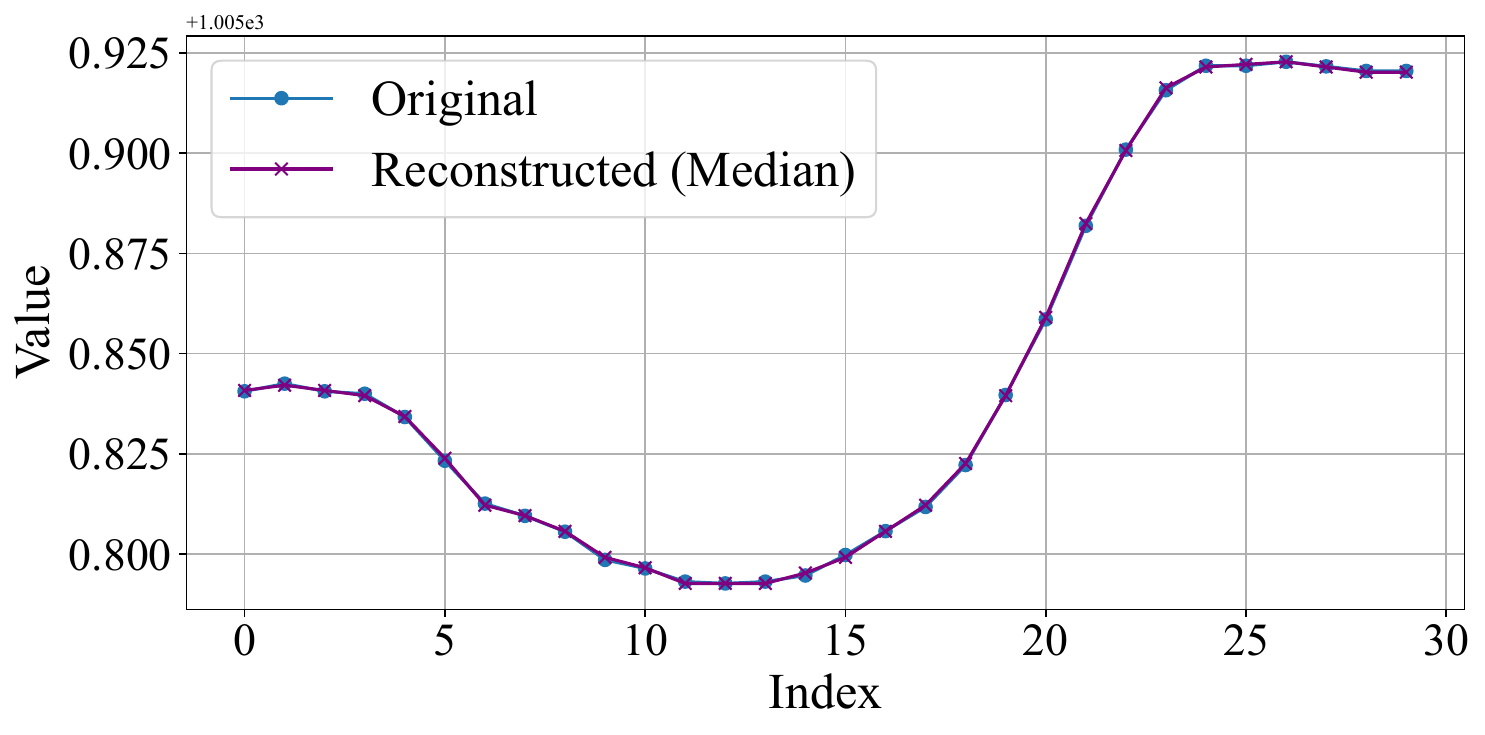}
\label{fig:r10}
\end{minipage}%
}%
\subfigure[$\alpha$ = 0.9, speed.]{
\begin{minipage}[t]{0.3\linewidth}
\centering
\includegraphics[width=\linewidth]{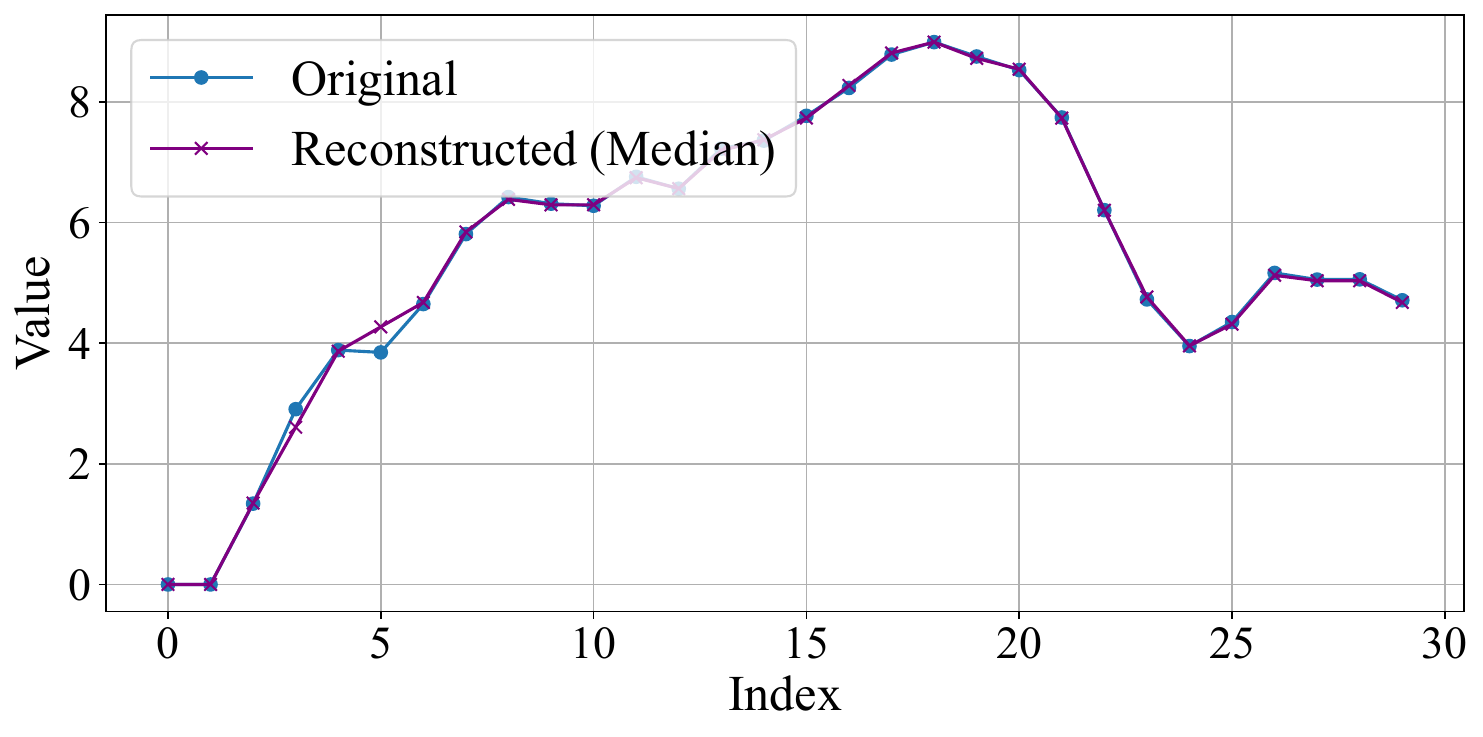}
\label{fig:r11}
\end{minipage}%
}%
\subfigure[$\alpha$ = 0.9, altitude.]{
\begin{minipage}[t]{0.3\linewidth}
\centering
\includegraphics[width=\linewidth]{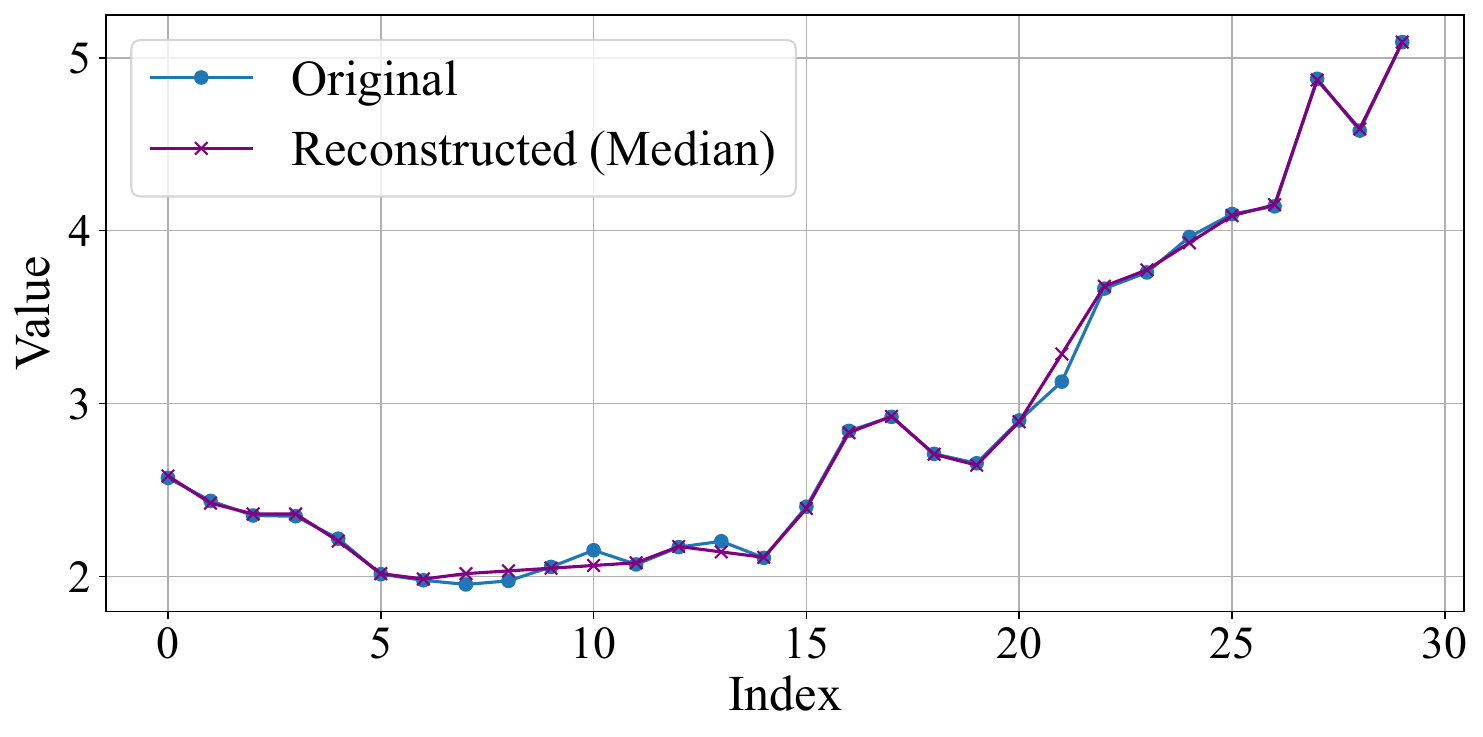}
\label{fig:r12}
\end{minipage}%
}%
\centering
\caption{\textbf{Reconstructed results.}}
\label{fig:result_example}
\vspace{-0.1in}
\end{figure*}

\textbf{Implementation.} We have chosen GPT-4~\cite{achiam2023gpt}, the most advanced and powerful LLM currently available, to conduct a thorough analysis of our datasets. Additionally, we assess the effectiveness of reconstructing three types of sensor data across various compression rates and transportation modes. This comparative analysis allows us to understand how different settings impact reconstruction accuracy.

\textbf{Metric.} For evaluating the accuracy of reconstructed sensor data, we employ the Mean Squared Error (MSE):
\[
\text{MSE} = \frac{1}{n} \sum_{i=1}^{n} (y_i - \hat{y}_i)^2.
\]
Here, \( n \) denotes the number of data points, \( y_i \) the actual values, and \( \hat{y}_i \) the predicted values. This metric highlights the average of the squared discrepancies between the predicted and actual data points, emphasizing larger errors, which is crucial for fine-tuning our reconstruction process.

\subsection{Overall Performance.}
Fig.~\ref{fig:eva} presents the accuracy percentages for sensor data reconstruction across taxi, MTR, and bus scenarios, utilizing the method implemented in \SystemName. For each scenario, the figure also reports the reconstruction accuracy at various compression ratios denoted by $\alpha$.
Specifically, the reconstruction accuracy achieved in taxi and bus settings was slightly lower than that in the MTR setting. This discrepancy can be attributed to the slower and more stable operation of the MTR compared to taxis and buses, which results in less variability in the sensor data and consequently reduces the difficulty of reconstruction.
For different compression ratios denoted by \(\alpha\), the reconstruction accuracy increases with higher values of \(\alpha\). Fig.~\ref{fig:result_example} illustrates this phenomenon, showing that as the compression ratio parameter \(\alpha\) is adjusted from 0.5 to 0.9, the reconstruction accuracy of the three types of sensor data improves visibly. This improvement is primarily because higher compression ratios retain more original data, reducing the information loss during the compression process and thus facilitating more accurate reconstruction.
The results demonstrate the high effectiveness of \SystemName across different scenarios.

\begin{figure*}
    \centering
    \includegraphics[width=0.8\linewidth]{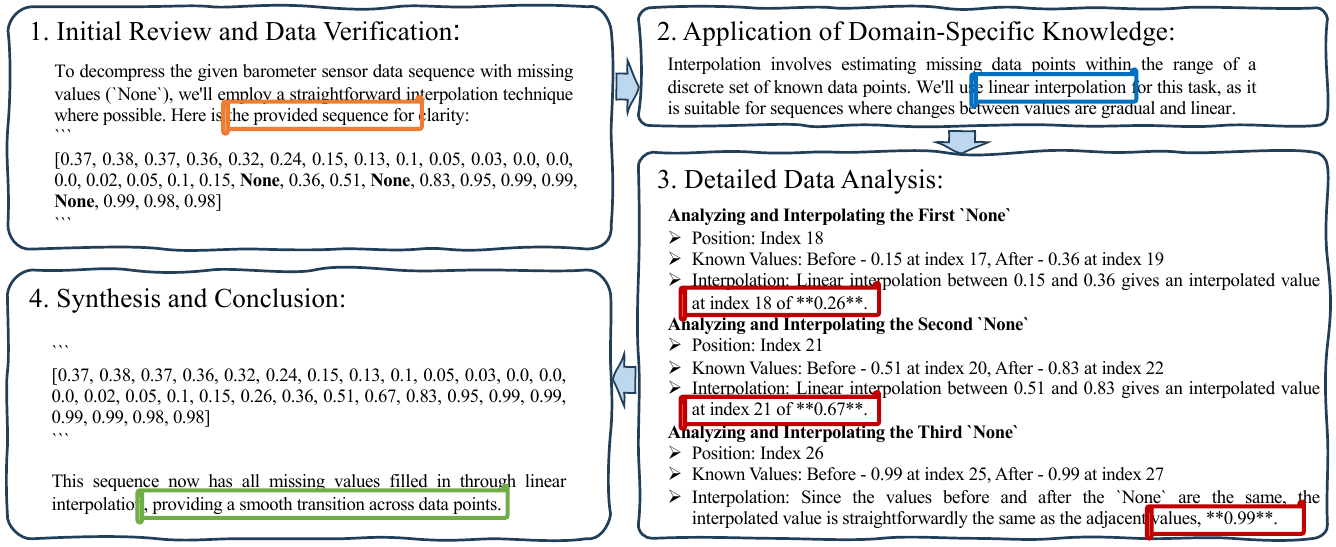}
    \caption{\textbf{Detailed step-by-step reconstruction of \SystemName.}}
    \label{fig:gpt4inference}
    \vspace{-0.1in}
\end{figure*}

\subsection{Detailed Inference Example.}
To illustrate the adeptness of LLMs in generating expert-level insights and precise inferences, we present a detailed example of how LLM reconstructs a sequence of air pressure sensor data in a taxi scenario, as depicted in Fig.~\ref{fig:gpt4inference}.
The inference methodology is divided into four distinct phases. 

The LLM starts by reviewing the dataset for completeness and identifying any missing data points. It then applies domain-specific knowledge to estimate the missing values using appropriate analytical methods. Next, the model performs a detailed analysis, interpolating values based on adjacent known data. Finally, it synthesizes the interpolated data, producing a reconstructed dataset with enhanced insights.
% \begin{itemize}
%     \item \textbf{Initial Review and Data Verification:} LLM begins by examining the dataset for completeness and identifying missing data points.
%     \item \textbf{Application of Domain-Specific Knowledge:} Utilizing embedded expertise, LLM applies suitable analytical methods to estimate missing values effectively.
%     \item \textbf{Detailed Data Analysis:} The model conducts a comprehensive analysis, calculating interpolated values by considering adjacent known data points.
%     \item \textbf{Synthesis and Conclusion:} After integrating the interpolated values, LLM synthesizes the updated data to deliver a reconstructed dataset with enhanced insights.
% \end{itemize}
This process exemplifies how LLMs can leverage specialized expertise to reconstruct transportation sensor data.

% \section{Discussion}

\section{Conclusion}
\label{sec:conclusion}
In this study, we demonstrate that LLMs can serve as a robust framework for accurately reconstructing compressed data from transportation sensors. This finding indicates that LLMs possess an inherent capability to interpret transportation sensor data autonomously, reducing the reliance on expert knowledge and manual effort.
Despite promising, it is crucial to pursue further research to define the precise conditions under which LLMs operate most effectively. By deepening our understanding of the strengths and limitations of LLMs, we can better leverage their capabilities to accurately reconstruct real-world sensor data, including wireless~\cite{ji2022sifall,ji2023construct,yang2023xgait,han2024seeing}, solar cell~\cite{ma2019solargest,wu2024xsolar}, biomedical ~\cite{xiao2021ulecgnet,duan2023emgsense}, and inertial~\cite{zhou2023one, xu2022washring} sensor data. This advancement will improve LLM deployment in smart sensing applications.

\vspace{-0.1in}
\section*{ACKNOWLEDGMENTS}
The work was supported by the Research Grants Council of the Hong Kong Special Administrative Region, China (Project No. CityU 21201420 and CityU 11201422), the Innovation and Technology Commission of Hong Kong (Project No. PRP/037/23FX and MHP/072/23).

\bibliographystyle{ACM-Reference-Format}
\bibliography{main}
\end{document}